\documentclass[letterpaper, 10 pt, conference]{ieeeconf}

\IEEEoverridecommandlockouts

\usepackage{etextools}
\usepackage{amssymb}
\usepackage{float}
\usepackage{makeidx}
\usepackage{amsmath}
\usepackage{bbm}
\usepackage{epsfig}
\usepackage{epsf}
\usepackage{psfrag}
\usepackage{verbatim}
\usepackage{color}
\usepackage{multirow}
\usepackage{tabularx}
\usepackage{booktabs}
\usepackage[tight,footnotesize]{subfigure}
\usepackage{array}
\usepackage{soul}
\usepackage{footnote}
\usepackage{lipsum}
\usepackage{cite}
\usepackage{dblfloatfix}

\usepackage{color, colortbl}
\usepackage[colorlinks,bookmarksnumbered,citecolor=orange,urlcolor=orange]{hyperref}
\usepackage{graphicx}
\graphicspath{{Figures}}
\DeclareGraphicsExtensions{.pdf,.png}
\usepackage{bigstrut}
\usepackage[english]{babel}

\usepackage{csquotes}

\usepackage[T1]{fontenc}
\usepackage{algorithm, setspace}
\usepackage{algpseudocode}
\usepackage{url}
\usepackage{multirow}
\usepackage{float}
\usepackage{xcolor}
\usepackage{hyperref}
 \hypersetup{
     colorlinks=true,
     linkcolor=orange,
     filecolor=orange,
     citecolor=orange,      
     urlcolor=orange,
     }

\usepackage{soul} 

\newcommand{\p}[1]{\smallskip \noindent \textbf{{#1}.}}

\usepackage{balance}

\title{\LARGE
Feeling Through the Load: Compliant Quadruped Locomotion under Payload Interactions
}

\author{Shaunak A. Mehta$^1$, Mayank Mishra$^2$, Prajit KrisshnaKumar$^1$, Sebastian Scherer$^2$ and Koichiro Niinuma$^1$
\thanks{$^{1}$ S. A. Mehta, P. KrisshnaKumar and K. Niinuma are with Fujitsu Research of America, Pittsburgh, PA, 15213.
\newline $^{2}$ M. Mishra and S. Scherer are with AirLab, Robotics Institute, Carnegie Melon University, Pittsburgh, PA, 15207.
\newline
corresponding author's e-mail: \texttt{mshaunak@fujitsu.com}}
\vspace{-1.5em}
}

\begin{document}
\maketitle
\begin{abstract}
Quadruped robots are increasingly expected to carry objects while moving through human environments. But what happens when a person interacts directly with the payload rather than with the robot? If the payload is unrestrained, the robot must distinguish intentional external interactions from ordinary payload motion, while still keeping the load balanced and maintaining stable locomotion. How can a quadruped infer and compliantly respond to such interactions using only onboard measurements? In this work, we develop a force-aware locomotion framework that treats payload interactions as commands that shape the motion of the combined robot-payload system. Our approach separates the learning of force-aware locomotion and force estimation on an unrestrained payload. We combine a compliant load-carrying policy with a causal force estimator, trained through estimator-in-the-loop data aggregation and finetuning, to predict interactions from onboard robot measurements. Our simulations and real-world experiments show that the resulting controller can maintain stable payload-carrying locomotion, yield compliantly to external interactions, and use the inferred force to support human-guided changes in the robot’s trajectory. Our project page is available at \href{https://feelthroughload.github.io/}{https://feelthroughload.github.io/}
\end{abstract}


\section{Introduction}

\begin{figure}[]
	\begin{center}
		\includegraphics[width=0.9\columnwidth]{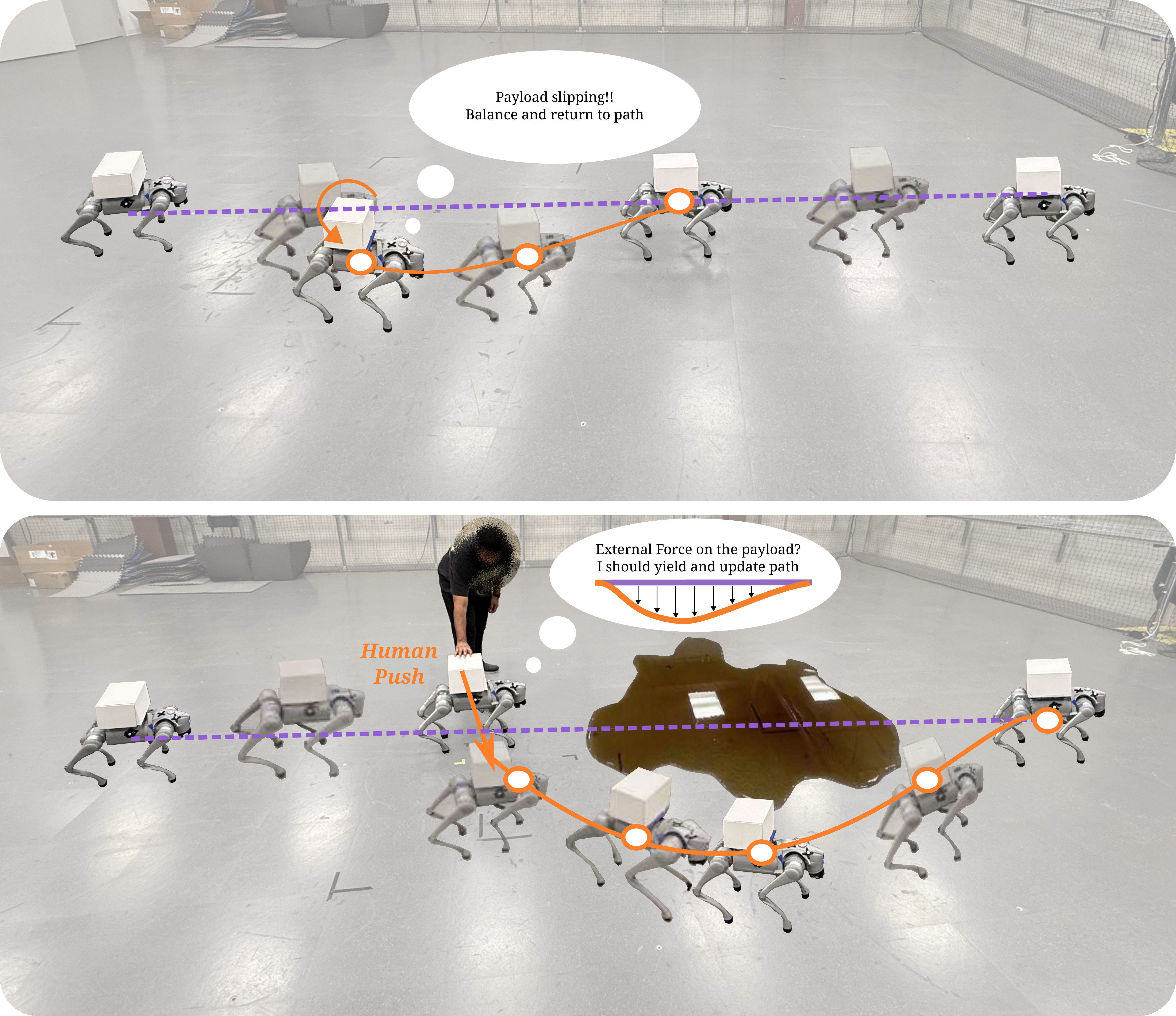}
		\vspace{-0.5em}
		\caption{Robot inferring and reacting to forces felt on an unrestrained payload. (Top) The robot detects a payload slip and tries to balance the payload, before returning to the planned path. (Bottom) The robot detects an external interaction on the payload, compliantly yields to the interaction, and updates its path to avoid a potential hazard.}
		\label{fig:front}
		\vspace{-2.0em}
	\end{center}
\end{figure}

Imagine a quadruped robot operating on a factory floor, transporting a payload from one station to another along a given trajectory. The robot, however, is unaware of a potential hazard in its path, such as a liquid spill. A human operator notices that the robot is moving toward the hazard and intervenes by pushing on the payload to guide it away from the hazardous path (see Figure~\ref{fig:front}). In such a situation, the robot not only needs to recognize an external interaction acting on a payload that is not rigidly attached to its body, but also needs to react safely and compliantly by yielding to the interaction or changing its path. At the same time, the robot should also adjust its motion to account for payload slip or imbalance caused due to the human push or movement during transportation.

Recent research in quadruped control has begun moving beyond locomotion over challenging terrains toward robots that can interact compliantly with people and adapt to the payloads they carry. Some works focus on enabling robots to
respond to forces acting on their bodies, either by rejecting them as disturbances or yielding to them for safer physical interaction~\cite{defazio2023seeing,xiao2024pa,xu2025facet}, while others adapt locomotion to the properties or motion of the carried payload~\cite{kurva2025mule,chang2025beyond}. More recent works on collaborative transportation use forces applied by a human or another robot to guide the robot's motion, but generally assume a mechanically coupled payload~\cite{turrisi2024pacc,cao2026paint}.

However, these assumptions may not hold in real-world transportation tasks. The payload may simply rest on the robot without being rigidly attached, its physical properties may be unknown, and adding sensors directly to the payload may not be practical. In such settings, the robot must rely on its own onboard measurements to infer the external interaction and determine how to respond. This leads us to the central question of this work:\\
\textbf{\textit{How can a quadruped infer and compliantly respond to external forces applied to an unrestrained payload?}}

In this paper, we address this question by treating the forces experienced on the payload as interactions that should influence the motion of the combined robot-payload system. 
Inspired by recent extensions of impedance control from robotic manipulators to legged locomotion \cite{xu2025facet,zhou2025hac,cao2026paint}, we use a virtual mass-spring-damper reference to describe how the robot should yield to these interactions. 
We learn a force-aware load bearing locomotion policy that realizes this compliant response while keeping the payload balanced, together with a causal estimator that recovers the applied force from onboard robot measurements. The explicit force estimate can also guide higher-level behavior, allowing a human to redirect the robot's trajectory by pushing directly on the payload.

Our work is most closely related to FACET \cite{xu2025facet} and prior work on collaborative load transportation \cite{cao2026paint,khandelwal2025compliant}. FACET considers compliant locomotion under external forces acting on the robot, while collaborative transportation methods enable robots to respond to interactions through a shared payload assuming that the payload is mechanically connected to the robot. In contrast, we consider a robot carrying an unrestrained payload and investigate how it can respond compliantly to forces applied directly to the payload while keeping it balanced.

Overall, we make the following contributions:

\p{Payload-Aware Compliant Locomotion}
We formulate the problem of compliant locomotion under forces applied directly to an unrestrained payload. In this formulation, the payload interaction shapes the desired motion of the combined robot-payload system while the robot maintains stable locomotion and keeps the payload balanced.

\p{Estimating and Responding to Payload Interactions}
We introduce a three-stage learning pipeline for force-aware load bearing locomotion. We first learn a privileged policy with recurrent adaptation, followed by causal payload force estimation using onboard sensors, and estimator-aware policy finetuning. Together, these stages enable the robot to explicitly estimate and respond to interactions with an unrestrained payload using only onboard robot measurements, without instrumenting the payload.

\p{Simulated and Real-World Experiments}
We evaluate our approach under diverse payload, interaction-force, and impedance conditions in simulations and on a real quadruped. We further demonstrate that the explicit force estimate provides a useful signal for downstream behavior, allowing a human to guide the robot's trajectory by interacting directly with the payload.

\section{Related Work} \label{sec:related}

\p{Compliant force adaptive locomotion}
Impedance and admittance control have been widely used in robotic manipulation to prescribe compliant motion in response to external forces~\cite{hogan1985impedance,ott2015hybrid,sun2024fixed}. These ideas have also influenced force-aware control for legged robots. Model-based methods incorporate compliant objectives and external disturbances into whole-body or optimization-based controllers~\cite{xin2018model,xin2020optimization}. Some other methods estimate external wrenches acting on the robot without dedicated force sensors using dynamics-based observers and state estimators~\cite{kang2023external,zhou2025simultaneous}.

Learning-based methods have similarly been used either to reject external disturbances or interpret them as interaction commands~\cite{xiao2024pa,chen2023learning}. The Seeing-Eye system jointly learns force-responsive locomotion and an onboard force estimator whose output guides navigation~\cite{defazio2023seeing}. However, the interaction is transmitted directly to the robot through a leash or handle. Standard velocity-tracking policies do not prescribe a response to external forces and can therefore behave stiffly during interaction~\cite{hartmann2024deep}. To tackle this, FACET learns to track the response of a virtual mass-spring-damper system~\cite{xu2025facet}, HAC-Loco estimates external disturbances from proprioception to adapt locomotion commands~\cite{zhou2025hac}, and SAC-Loco combines adjustable compliance with a learned safety and recovery mechanism~\cite{zhang2025sac}. Force-conditioned policies have also been developed for legged loco-manipulation, where estimated or commanded forces regulate an end effector or modify its motion target~\cite{portela2024learning,zhi2025learning}. In these settings, the interaction acts directly on the robot body or end effector. In contrast, we consider forces applied directly to a payload that is coupled to the robot only through frictional contact.

\p{Payload-Aware Locomotion and Interaction}
Prior work on payload-carrying quadrupeds has primarily considered how locomotion should adapt to the carried load. Model-based methods estimate payload properties such as mass and center-of-mass offsets online~\cite{tournois2017online}, while learning-based approaches adapt locomotion to unknown or dynamically varying loads~\cite{kurva2025mule}. Beyond Robustness estimates the motion characteristics of an unrestrained payload from proprioceptive history and adapts locomotion to stabilize it~\cite{chang2025beyond}. 
These methods address payload adaptation and stabilization, but do not consider an external agent intentionally interacting with an unrestrained payload to guide the robot.

Collaborative transportation instead uses interaction through a shared payload to infer partner intent.  Passive impedance arms convert payload-induced deflections into guidance commands~\cite{turrisi2024pacc}, while suction-cup end effectors measure interaction wrenches for admittance-based transport~\cite{plotas2025control}. External base wrenches have also been estimated for cooperative carrying through rigidly connected loads~\cite{khandelwal2025compliant}. PAINT infers partner interaction wrenches from proprioceptive histories by exploiting the coupled dynamics of a shared payload~\cite{cao2026paint}, and related human-humanoid work learns proprioception-only compliant coordination during shared-object carrying~\cite{du2025learning}. 
However, these methods rely on a mechanically coupled object as the interaction interface.

Together, these works address force-aware interaction, payload stabilization, and collaborative transportation, but treat them as separate problems.
In contrast, we consider forces applied directly to an unrestrained payload that is coupled to the robot only through frictional contact. Our work addresses these jointly by estimating the payload interaction from onboard measurements and using it for compliant locomotion while maintaining payload stability.

\section{Problem Statement} \label{sec:problem}

We consider a quadruped carrying an unrestrained payload on a trunk-mounted support. Because the payload is not rigidly connected, it may move relative to the robot or leave the support. Returning to our motivating example, an external agent may apply a force $\mathbf{F}_{L,t}$ directly to the payload, and the desired behavior is a prescribed compliant response that retains the payload.

Inspired by impedance control in robotic manipulation, we define the desired force--motion response through a virtual mass-spring-damper reference~\cite{ott2015hybrid,sun2024fixed, xu2025facet}.
Let $x^d$ denote the nominal motion command and $x^r$ be the compliant reference. The desired response to an external payload interaction can be defined as:
\begin{equation}
    {M}_{v}\ddot{x}^{r}
    ={K}_{p}(x^{d}-x^{r})
     +{K}_{d}(\dot{x}^{d}-\dot{x}^{r})
     +{F}_{L,t}
    \label{eq:problem_compliant_reference}
\end{equation}
where ${M}_{v}$, ${K}_{p}$, and ${K}_{d}$ specify the commanded virtual mass, stiffness, and damping. Without an external interaction, the reference converges to the nominal motion, while a force deflects it according to the commanded impedance. Our goal is to realize this response by estimating payload force while maintaining locomotion stability and payload retention.

Let $s_t \in \mathcal{S}$ be the true state of the system
and $a_t \in \mathcal{A}$ be the robot action commanding the desired joint position offsets. During deployment, a controller $\Pi$ only receives the observation $o_t \in \mathcal{O}$, comprising the commanded motion and on-board robot measurements, 
and selects an action $a_t$. The objective of this controller is to maximize the reward function $r(s_t, a_t)$ over an episode of horizon $T$.
In order to realize this objective, the robot should first be able to infer the direction and magnitude of the payload interaction. 
However, since the force acts on the payload, its effects reach the robot only through uncertain frictional payload–support contact. 
Additionally, the resulting implicit signals depend on the payload mass and center of mass (CoM), the current robot gait, and the ground-contact configuration. The robot should therefore be able to distinguish the interaction-induced motion from the inertial and ground-reaction effects of ordinary locomotion. To tackle this challenge, we aim to learn an explicit causal estimate of the external force ($\widehat{\mathbf{F}}_{L,t} =f_{\phi}(o_{\leq t})$)
that remains informative using only information available onboard the robot.

In this problem, force estimation and control are coupled in a closed loop. The force estimate deforms the compliant reference and changes the robot's response, which consequently changes the observations used to produce subsequent estimates. The estimator must therefore remain informative under the state distribution induced by its own predictions. This feedback also exposes a physical trade-off between compliance and stability. Excessive rigidity would reject the intended interaction, while excessive yielding can lead to locomotion instability or loss of payload.

Beyond immediate compliant locomotion, an explicitly inferred force provides a physically meaningful command signal for higher-level behavior. As we will show in our hazard avoidance experiment in Section \ref{sec:simulation_deflection}, it is used to deform a nominal path in the direction of the interaction. We treat this trajectory update as a downstream application of the force-aware controller rather than part of the core force estimation problem.

\section{Method} \label{sec:method}

\begin{figure*}[]
	\begin{center}
		\includegraphics[width=1.8\columnwidth]{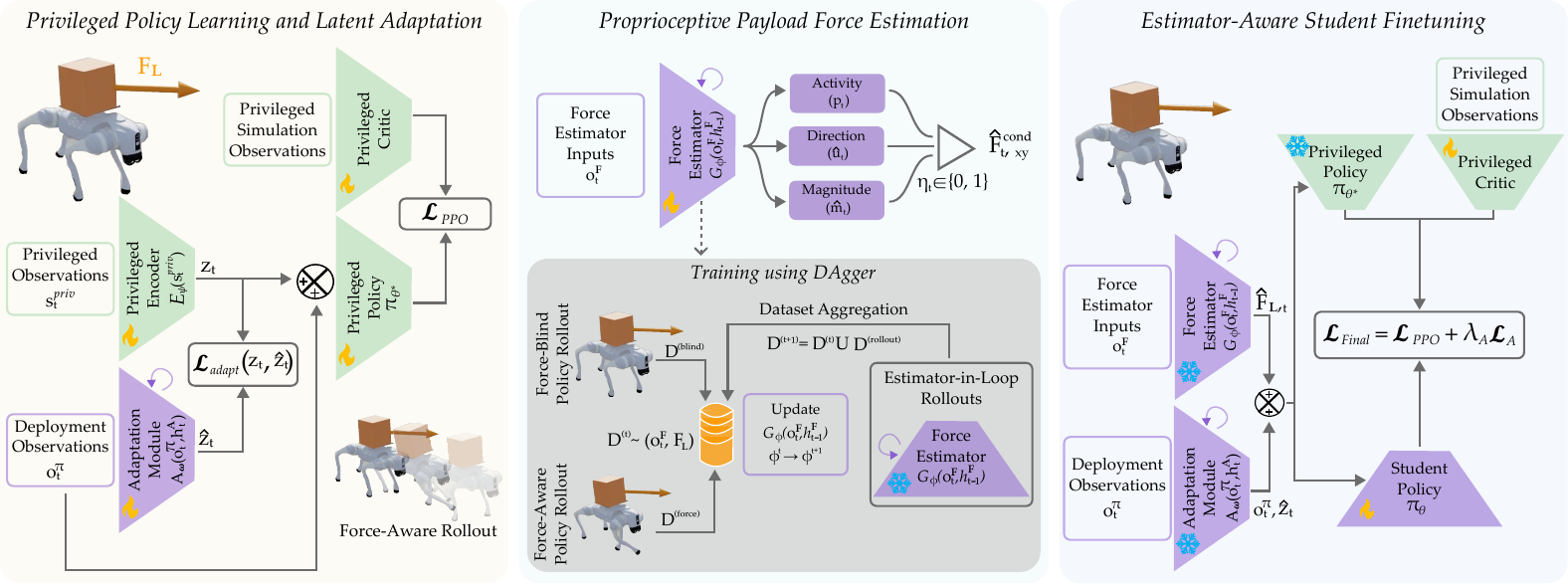}
		\vspace{-0.5em}
		\caption{Overview of our proposed approach for compliant locomotion under payload interactions. (Left) Privileged policy learning trains a force-aware locomotion policy, while a recurrent adaptation module learns to recover its privileged latent from deployable observations. (Center) A causal payload force estimator predicts interaction activity, direction, and magnitude from onboard  measurements and is trained through mixed-distribution data aggregation. (Right) The estimator and adaptation module are frozen while the actor is finetuned under estimator-driven rollouts using PPO and behavioral anchoring toward the frozen oracle force-adapted policy (privileged policy). Green components require privileged or true force inputs, whereas purple components form the deployable inference path.}
		\label{fig:methods}
		\vspace{-2.1em}
	\end{center}
\end{figure*}

In this section, we introduce a three-stage training pipeline for deployable force-conditioned locomotion (see Figure \ref{fig:methods}). We first learn the desired force-aware behavior using privileged state and the true payload force, while training a recurrent adaptation module to recover the privileged latent from deployable observations. We then replace the true force input with a causal estimator learned from onboard robot measurements. Finally, we finetune the policy under estimator-driven rollouts to account for errors introduced by the learned force estimate.

\subsection{External Interaction Aware Compliant Locomotion}
\label{sec:force_conditioned_policy}

To generate the desired compliant motion, we use the virtual mass-spring-damper reference defined in Eq. \ref{eq:problem_compliant_reference}, following the formulation introduced by FACET~\cite{xu2025facet}, and extend it to interaction applied directly to the payload.  
We maintain synchronized reference states anchored to the trunk and payload, both driven by the same nominal setpoint, commanded impedance, and payload-interaction force.
During privileged policy learning and recurrent adaptation, the reference generator receives the true payload force, $\mathbf{F}_{c}=\mathbf{F}_{L}$, while during estimator-aware finetuning and deployment, it is replaced by the causal estimate ($\mathbf{F}_{c}=\widehat{\mathbf{F}}_{L}$). The same force is supplied directly to the locomotion policy, while the reference dynamics generate position, velocity, and yaw targets over a short horizon.

\p{Compliance and Payload-Retention Objective}
The compliant reference specifies how the robot should yield to a payload interaction. 
The FACET objective rewards compliant-reference tracking and stable locomotion, but does not account for retention of an unrestrained payload \cite{xu2025facet}. We therefore augment it with terms that regulate the payload position and motion relative to its support.

Let $\mathbf{d}^{B}_t=[d_{x,t},d_{y,t},d_{z,t}]^\top$ denote the payload position relative to the center of the support, and let $(w_x,w_y)$ denote the support half-widths. We define the normalized horizontal offset and its associated reward as
\begin{equation}
    \rho_t^2 = (\frac{d_{x,t}}{w_x})^2 + (\frac{d_{y,t}}{w_y})^2
    \quad
    r^{\mathrm{margin}}_t = \exp(-\frac{\rho_t^2}{\sigma_m^2})
    \label{eq:payload_margin_reward}
\end{equation}
where $\sigma_m$ controls sensitivity to displacement. The normalization accounts for the support dimensions, while the exponential provides a smooth penalty near its boundary.

However, position alone does not capture payload motion toward an edge. Let $\mathbf{v}^{\mathrm{rel}}_{L,t,xy}$ denote payload velocity relative to the trunk. We penalize this motion and provide a retention bonus while the payload remains within the support bounds:
\begin{align}
    r^{\mathrm{slip}}_t &= -\|\mathbf{v}^{\mathrm{rel}}_{L,t,xy} \|_2^2, \label{eq:payload_slip_reward}\\
    r^{\mathrm{retain}}_t &= \mathbb{I}[d_{z,t}>z_{\min} \ \wedge\  \|\mathbf{d}^{B}_{t,xy}\|_2<d_{\max}]
    \label{eq:payload_retention_reward}
\end{align}
where $z_{\min}$ and $d_{\max}$ define the retention bounds. We additionally encourage a level support using $r^{\mathrm{level}}_t=1-\|\mathbf{g}^{B}_{t,xy}\|_2^2$. Denoting the unchanged FACET objective by $r^{\mathrm{FACET}}_t$, the complete reward for our objective can be given as:
\begin{equation}
    r_t = r^{\mathrm{FACET}}_t
    +\alpha_m r^{\mathrm{margin}}_t
    +\alpha_s r^{\mathrm{slip}}_t
    +\alpha_r r^{\mathrm{retain}}_t
    +\alpha_u r^{\mathrm{level}}_t
    \label{eq:complete_reward}
\end{equation}
where the coefficients $\alpha_m$, $\alpha_s$, $\alpha_r$, and $\alpha_u$ control the balance between compliant locomotion and payload stability.

\p{Policy Observations and Privileged Policy Learning}
Now that we have defined 
an objective that can be used to evaluate the behavior, we move to learning a policy that maps the task-level objective into joint commands using onboard robot measurements. In our framework, we provide the impedance command, robot-side measurements, and force input to the policy as
\begin{align}
    \mathbf{o}^{\pi}_t =
    [&
        \mathbf{c}_t,
        \mathbf{g}^{B}_{t-H_o:t},
        \mathbf{q}_{t-H_o:t},
        \dot{\mathbf{q}}_{t-H_o:t},
        \nonumber \\
        &\mathbf{a}_{t-H_a:t-1},
        \mathbf{f}^{\mathrm{imu}}_t,
        \boldsymbol{\tau}^{\mathrm{est}}_t,
        \mathbf{F}^{B}_{c,t}
    ]
    \label{eq:policy_observation_compact}
\end{align}
where $\mathbf{c}_t$ contains the impedance command and setpoint error, $\mathbf{g}^{B}$ is projected gravity, and $\mathbf{q}$ and
$\dot{\mathbf{q}}$ are the joint positions and velocities. The policy also receives previous actions, trunk IMU and actuator-torque measurements collected within the current control interval. Here, $H_o$ and $H_a$ denote the observation and action history horizons. The final three dimensions form the explicit external force channel.

During the first training stage, we leverage these observations along with privileged simulation-only information to learn the desired force-aware behavior. We use an asymmetric actor-critic formulation in which the actor receives $\mathbf{o}^{\pi}_t$ together with a compact privileged latent state:
\begin{equation}
    \mathbf{z}^{\mathrm{priv}}_t
    =E_{\psi}(\mathbf{s}^{\mathrm{priv}}_t),
    \mathbf{a}_t\sim
    \pi_{\theta}(
        \cdot\mid
        \mathbf{o}^{\pi}_t,
        \mathbf{z}^{\mathrm{priv}}_t
    )
    \label{eq:privileged_policy}
\end{equation}
The privileged state $\mathbf{s}^{\mathrm{priv}}_t$ contains reference, robot, payload, payload-properties, and contact information. The payload-interaction force is excluded because it is already provided through the force channel in $\mathbf{o}^{\pi}_t$. The critic receives the policy observation together with the uncompressed privileged state. We jointly optimize the actor $\pi_{\theta}$, privileged encoder $E_{\psi}$, and critic with PPO~\cite{schulman2017proximal} using the payload-aware objective introduced in Eq. \ref{eq:complete_reward}.

\p{Recurrent Latent Adaptation}
While privileged latent helps the policy learn the desired behavior, it cannot be constructed outside simulation. We train a recurrent adaptation module to recover this latent from the policy observations \cite{kumar2021rma}:
\begin{equation}
    (
        \widehat{\mathbf{z}}^{\mathrm{priv}}_t,
        \mathbf{h}^{A}_{t+1}
    )
    =
    A_{\omega}(
        \mathbf{o}^{\pi}_t,
        \mathbf{h}^{A}_t
    )
    \label{eq:recurrent_privileged_adaptation}
\end{equation}
where $\mathbf{h}^{A}_t$ carries a learned summary across control steps, complementing the fixed observation histories contained in $\mathbf{o}^{\pi}_t$. Using rollouts generated by the privileged policy, we initially optimize
\begin{equation}
    \mathcal{L}_{\mathrm{adapt}}
    =
    \mathbb{E}_{t}[
        \|
            \widehat{\mathbf{z}}^{\mathrm{priv}}_t
            -
            \mathbf{z}^{\mathrm{priv}}_t
        \|_2^2
    ]
    \label{eq:adaptation_objective}
\end{equation}
The privileged latent is treated as a fixed target, so this loss updates only $A_{\omega}$. During this stage, the actor continues to receive
$\mathbf{z}^{\mathrm{priv}}_t$; consequently, errors in $\widehat{\mathbf{z}}^{\mathrm{priv}}_t$ do not affect the collected trajectories.

When the recurrent estimate replaces the privileged latent at the actor input, its errors change the actor outputs and the resulting state distribution. To expose the adaptation module to this closed-loop distribution, we freeze the trained actor and generate rollouts using
$\mathbf{a}_t\sim \pi_{\theta^\star}(\cdot\mid\mathbf{o}^{\pi}_t, \widehat{\mathbf{z}}^{\mathrm{priv}}_t)$.
We continue minimizing Eq.~\eqref{eq:adaptation_objective}, retaining simulator-only information solely to construct the target latent. The actor and privileged encoder remain fixed, and only $A_{\omega}$ is updated during this stage.
The resulting adaptation module removes the policy dependency on privileged state during inference, but the actor still needs the true payload force through $\mathbf{o}^{\pi}_t$. We refer to it as the \emph{oracle force-adapted policy} ($\pi_{\theta^*}$).

\subsection{Proprioceptive Payload Force Estimation}
The recurrent adaptation module removes the need for privileged state, but the oracle force-adapted policy still requires the true payload force for both the compliant reference and policy force channel. We remove this remaining oracle input using a causal estimator that predicts the payload interaction from onboard robot measurements.

\p{Estimator Architecture}
Let $\mathbf{o}^{\pi,-F}_t$ denote the policy observation in Eq.~\ref{eq:policy_observation_compact} with its force channel removed. We augment these measurements with the normal foot loads $\boldsymbol{\ell}^{\mathrm{foot}}_t$ and corresponding contact states $\boldsymbol{\chi}^{\mathrm{foot}}_t$ to get the input for the force estimator as:
\begin{equation}
    \mathbf{o}^{F}_t = [\mathbf{o}^{\pi,-F}_t,\boldsymbol{\ell}^{\mathrm{foot}}_t,\boldsymbol{\chi}^{\mathrm{foot}}_t]
    \label{eq:estimator_observation}
\end{equation}

The estimator therefore receives robot-side proprioceptive, inertial, actuator, and foot-load measurements, but not the policy force channel,
ground-truth force, payload pose, or payload properties. A recurrent encoder summarizes these measurements as $\mathbf{h}^{F}_t = G_{\phi}(\mathbf{o}^{F}_t, \mathbf{h}^{F}_{t-1})$.

Instead of directly predicting the force vector using the recurrent model, we leverage the predicted hidden state to predict the interaction probability, force direction, and magnitude.Let $g_p$, $g_u$, and $g_m$ denote the probability, direction, and magnitude heads, respectively. These heads map the recurrent state $\mathbf{h}^{F}_t$ to their corresponding outputs:
\begin{align}
    \widehat{p}_t =\operatorname{sigmoid}(g_p&(\mathbf{h}^{F}_t)),
    \widehat{m}_t =s_F\operatorname{softplus} g_m(\mathbf{h}^{F}_t)) \nonumber \\
    \widehat{\mathbf{u}}_t &=\frac{g_u(\mathbf{h}^{F}_t)} {\|g_u(\mathbf{h}^{F}_t)\|_2+\epsilon}
    \label{eq:decomposed_force_estimate}
\end{align}
where $\widehat{p}_t$ estimates the probability that the applied force exceeds the minimum actionable-force threshold $F_{\min}$, $\widehat{\mathbf{u}}_t$ is the predicted unit direction, and $\widehat{m}_t$ is the predicted force magnitude. Their conditional force prediction can be given as $\widehat{\mathbf{F}}^{\mathrm{cond}}_{t,xy} =\widehat{m}_t\widehat{\mathbf{u}}_t$. The constant $\epsilon$ ensures numerical stability during normalization, while $s_F$ scales the magnitude output to the force range used during training.

At inference, we obtain a binary activity state from the predicted probability using a causal hysteresis gate with separate onset and release thresholds:
\begin{equation}
    \eta_t=\Gamma(\widehat{p}_{\leq t}),
    \quad
    \widehat{\mathbf{F}}^{B}_{L,t} = \eta_t
    \begin{bmatrix}
        \widehat{\mathbf{F}}^{\mathrm{cond}}_{t,xy}\\
        0
    \end{bmatrix},
    \quad
    \eta_t\in\{0,1\}
    \label{eq:deployable_force_estimate}
\end{equation}
where $\Gamma$ denotes the causal hysteresis gate. This suppresses force predictions when no actionable interaction is detected without attenuating the magnitude of active predictions. The resulting $\widehat{\mathbf{F}}^{B}_{L,t}$ is supplied to both the compliant reference and the policy force channel.

\p{Training Objective}
We label a sample as an actionable interaction using $y_t=\mathbb{I}[\|\mathbf{F}^{B}_{L,t,xy}\|_2>F_{\min}]$. For active samples, we define $m_t=\|\mathbf{F}^{B}_{L,t,xy}\|_2$ and $\mathbf{u}_t=\mathbf{F}^{B}_{L,t,xy}/m_t$, respectively. Because force direction and magnitude are not meaningful without an actionable interaction, we supervise them only on active samples:
\begin{align}
    \mathcal{L}_{F} &= \beta_{\mathrm{prob}}\mathcal{L}^{\mathrm{bal}}_{\mathrm{BCE}} (\widehat{p}_t,y_t) + \beta_{\mathrm{dir}} \mathbb{E}_{y_t=1} [1-\widehat{\mathbf{u}}_t^\top\mathbf{u}_t] \nonumber\\
    &+ \beta_{\mathrm{mag}} \mathbb{E}_{y_t=1} [\mathcal{H}(\widehat{m}_t-m_t)] \nonumber \\
    &+ \beta_{\mathrm{vec}} \mathbb{E}_{y_t=1}[\mathcal{H}(\widehat{\mathbf{F}}^{\mathrm{cond}}_{t,xy} - \mathbf{F}^{B}_{L,t,xy})]
    \nonumber\\
    &+ \beta_{0}\mathbb{E}_{y_t=0}[(\widehat{p}_t\widehat{m}_t)^2]
    \label{eq:force_estimation_loss}
\end{align}
Here, $\mathcal{H}$ denotes the Huber loss. The balanced binary cross-entropy accounts for the imbalance between active and inactive samples,
while the final term suppresses confident nonzero predictions during ordinary locomotion.
We additionally predict the payload mass through a training-only head $\widehat{M}_{L,t}=g_M(\mathbf{h}^{F}_t)$ and optimize $\mathcal{L}_{\mathrm{est}} = \mathcal{L}_{F} + \beta_M \mathbb{E}_t[\mathcal{H}(\widehat{M}_{L,t}-M_L)]$. This auxiliary loss encourages the estimator to distinguish changes in payload inertia from external interaction. 

\p{Mixed-Distribution DAgger}
An estimator trained only on oracle-force rollouts may not generalize to the states induced by its own predictions. We therefore train the initial estimator on a mixture of blind and oracle-force trajectories. In blind trajectories, the interaction acts physically on the payload but is withheld from the reference and policy, while in oracle trajectories, the true force drives both.

Following a DAgger-style procedure~\cite{ross2011reduction}, we use the initial estimator to collect additional estimator-in-the-loop trajectories. During these rollouts, the estimator prediction drives the reference and policy, while the true force is retained only as a supervision label. We then retrain the final estimator on the union of the blind, oracle-force, and estimator-in-the-loop data. This aggregation may be repeated with successive estimators. In our experiments, we perform one round of estimator-in-the-loop data aggregation.

\subsection{Estimator-Aware Policy Finetuning}
\label{sec:estimator_aware_finetuning}

The learned force estimator removes the dependency on the true force input, but introduces magnitude, direction, gating, and timing errors that were absent during oracle force adaptation. We therefore initialize the actor from the oracle force-adapted policy, $\theta\leftarrow\theta^\star$, and finetune it with PPO under estimator-driven rollouts. During these rollouts, the estimated force drives both the compliant reference and the policy force channel. The force estimator and recurrent adaptation module remain frozen, while the actor and critic are updated. As in privileged policy learning, the critic receives the policy observation together with the uncompressed privileged state.

While PPO finetuning can improve robustness to estimation errors by making the actor less responsive to its force input, it may also weaken the desired compliant behavior. We therefore retain a frozen copy of the oracle force-adapted actor, $\pi_{\theta^\star}$, and regularize the finetuned actor toward its action distribution. We define this behavioral anchoring using an anchor loss given as:

\begin{equation}
    \mathcal{L}_{A}
    =
    \frac{\mathbb{E}_{t}[w(y_t) D_{\mathrm{KL}}\!\left(\pi_{\theta^*}\,\middle\|\, \pi_\theta\right)]
    }
    {\mathbb{E}_{t}[w(y_t)]}
    \label{eq:behavioral_anchor}
\end{equation}

where $\widetilde{\mathbf{o}}^{\pi}_t$ denotes the policy observation containing the estimated force, and $w(y_t)$ assigns different anchoring weights to active and inactive interactions. Both actors receive the same estimated-force observation and the same latent produced by the frozen recurrent adaptation module. The true force is used only to determine the training-time label $y_t$ and is not provided to either actor.
\begin{equation}
    \mathcal{L}_{\mathrm{final}} = \mathcal{L}_{\mathrm{PPO}} + \lambda_A\mathcal{L}_{A}
    \label{eq:estimator_aware_objective}
\end{equation}

where $\lambda_A$ controls the behavioral-anchor strength. 
The resulting policy is optimized under the estimator-induced state distribution while remaining close to the compliant behavior learned with oracle force. At deployment, all simulator-only quantities used for supervision, data aggregation, and anchor weighting are removed; the controller uses only the observations in Eqs.~\ref{eq:policy_observation_compact} and~\ref{eq:estimator_observation}.

\section{Simulations}
\label{sec:sims}

In this section, we evaluate whether the proposed controller realizes a commanded compliant response to payload interactions and whether its force estimate can guide a downstream trajectory correction.

\p{Experimental Setup}
We simulate a Unitree Go2 carrying a rigid, unrestrained payload on a trunk-mounted support. The $0.20$~m-tall, bottom-heavy payload has footprints ranging from $0.18\times0.18$~m to $0.35\times0.25$~m. Horizontal forces are applied near its geometric center with randomized longitudinal and lateral offsets. Payload and contact properties, robot dynamics, actuator properties, and initial conditions are randomized during training and evaluation.

\subsection{Commanded Compliance and Force Response}
\label{sec:c1c2}

\p{Independent Variables}
We compare \textbf{Ours} with the original \textsc{Facet} policy~\cite{xu2025facet}, a task-matched variant trained with our payload-retention objective ({\textsc{Facet}-L), the oracle-force adapted policy (\textbf{Oracle}), and the same force-conditioned policy driven by a filtered centroidal inverse-dynamics estimate (\textbf{Ours-Analytic}). Oracle receives the true payload force, whereas Ours uses the learned estimate. Ours-Analytic is a simulation-only baseline with access to simulator dynamics and payload mass.

We evaluate $m_L\in\{2,4,6\}$~kg, $F\in\{6,8,10,13,16,20,25,30\}$~N, and $K_p\in\{6,8,16,30\}$~N/m, with $K_d=1.8\sqrt{K_p}$~N\,s/m and $M_v=4$~kg. Each force is applied for two seconds, followed by a three-second recovery period. Every condition is tested in 32 randomized simulation instances.

\p{Dependent Variables}
The true-force impedance reference provides a common target for every method. We measure the response magnitude using the commanded compliance ratio $\rho_{\mathrm{cmd}}=\Delta x_B^{\parallel}/\Delta x_{\mathrm{ref}}^{\parallel}$, where $\Delta x_B$ and $\Delta x_{\mathrm{ref}}$ are the robot and reference displacements, and a value of one denotes the commanded response. Directional accuracy is measured by $e_\theta=\operatorname{atan2}(|\Delta x_B^{\perp}|,\Delta x_B^{\parallel})$, and termination is the fraction of trials ending because of a robot or payload failure. Compliance metrics are averaged over conditions in which Oracle succeeds in at least $95\%$ of trials; this condition set is then held fixed across methods, while termination is reported over the full grid.

\begin{table}[t]
    \centering
    \caption{Aggregate compliance response and full-grid termination.
    Values in the columns represent the mean and std dev across five seeds.}
    \vspace{-0.5em}
    \label{tab:simulation_compliance_aggregate}
    \scriptsize
    \setlength{\tabcolsep}{5pt}
    \begin{tabular}{l c c c}
        \toprule
        Method
        & $\rho_{\mathrm{cmd}}$ $\uparrow$
        & $e_{\theta}$ (deg) $\downarrow$
        & Termination (\%) $\downarrow$ \\
        \midrule
        Oracle           & $0.865\pm0.001$ & $1.77\pm0.04$ & $19.13\pm0.31$ \\
        \textsc{Facet}   & $0.252\pm0.002$ & $8.99\pm0.21$ & $52.91\pm0.23$ \\
        \textsc{Facet-L} & $0.702\pm0.002$ & $7.27\pm0.16$ & $21.57\pm0.08$ \\
        Ours-Analytic    & $0.602\pm0.003$ & $3.37\pm0.15$ & $33.48\pm0.44$ \\
        Ours             & $0.826\pm0.003$ & $3.73\pm0.15$ & $19.84\pm0.24$ \\
        \bottomrule
    \end{tabular}
    \vspace{-0.5em}
\end{table}

\begin{figure}[t]
    \centering
    \includegraphics[width=\columnwidth]
    {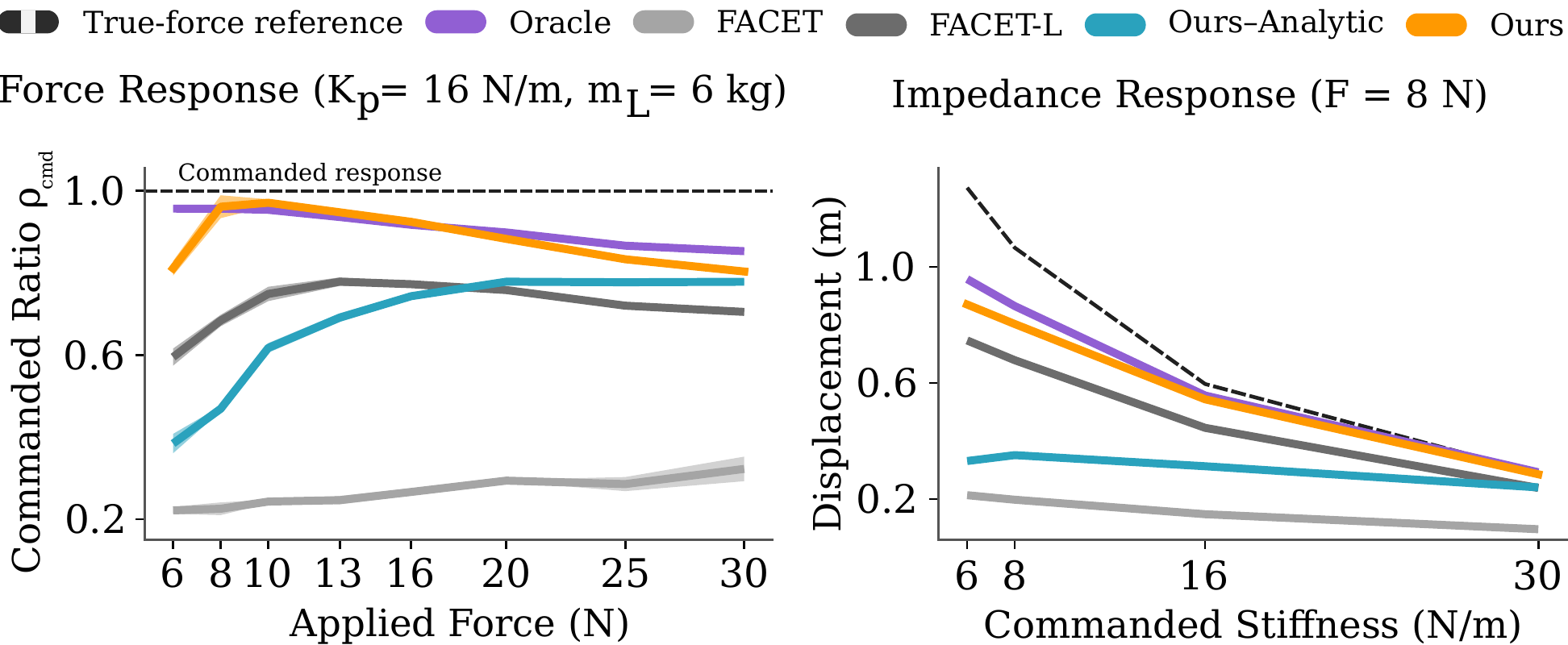}
    \caption{Compliance across diverse force and impedance settings. (left) Commanded compliance ratio at $K_p=16$ N/m and $m_L=6$ kg. (right) Displacement under an $8$ N force as commanded stiffness varies. Curves represent the means and shaded regions show standard deviations across five evaluation seeds.}
    \label{fig:sims}
    \vspace{-1.5em}
\end{figure}

\begin{table}[t]
    \centering
    \caption{
    Quantitative-analysis of the force estimator.
    Values in the columns represent the mean and std dev across five seeds}
    \vspace{-0.5em}
    \label{tab:simulation_force_estimation}
    \scriptsize
    \setlength{\tabcolsep}{3.2pt}
    \begin{tabular}{l c c c c}
        \toprule
        Force input
        & \shortstack{Magnitude\\ratio $\uparrow$}
        & \shortstack{Direction\\error (deg) $\downarrow$}
        & \shortstack{Detection\\(\%) $\uparrow$}
        & \shortstack{Onset\\(s) $\downarrow$} \\
        \midrule
        Learned  & $0.886\pm0.003$ & $10.87\pm0.34$ & $98.90\pm0.29$ & $0.175\pm0.001$ \\
        Analytic & $0.445\pm0.002$ & $14.24\pm0.47$ & $86.84\pm0.44$ & $0.433\pm0.006$ \\
        \bottomrule
    \end{tabular}
    \vspace{-2.0em}
\end{table}

\p{Results}
Table~\ref{tab:simulation_compliance_aggregate} shows that \textbf{Ours }approaches the \textbf{Oracle} response while improving substantially over both \textsc{Facet} variants. It reaches $\rho_{\mathrm{cmd}}=0.826\pm0.003$, compared with $0.702\pm0.002$ for \textsc{Facet-L}, while approximately halving its direction error. Its full-grid termination rate remains close to \textbf{Oracle}, where the largest failure rates occur under high forces on light payloads.

The estimator comparison in Table~\ref{tab:simulation_force_estimation} helps explain this improvement. Relative to the analytic estimate, the learned estimator recovers more of the applied force, detects interactions more consistently, and responds with less delay. Figure~\ref{fig:sims} further shows that \textbf{Ours} follows the \textbf{Oracle} trend as force and stiffness vary --- larger forces produce the corresponding compliant response, while increasing $K_p$ reduces displacement. The controller therefore realizes a tunable impedance rather than a fixed reaction to force detection.

\subsection{Force-Guided Trajectory Correction}
\label{sec:simulation_deflection}

We next test whether the explicit force estimate can serve as a command for downstream behavior by using an interaction to guide the robot around an obstacle.

\p{Independent Variables}
We compare \textbf{Ours} with \textbf{Oracle}, \textsc{Facet-L}, and \textbf{Ours-Analytic} under $F\in\{10,13,16,20\}$~N, $m_L\in\{2,4,6\}$~kg, and both lateral interaction directions. Every force-aware method uses the same trajectory correction rule but a different force input. Because \textsc{Facet-L} does not produce an explicit force estimate, its nominal path remains unchanged.

\p{Procedure}
The robot initially follows a straight route toward a goal with an obstacle in its path. An interaction is applied before the obstacle, causing the robot to pause and yield compliantly. After the interaction, the signed lateral impulse is mapped to a bounded minimum-jerk path deformation whose direction selects the avoidance side and whose finite support returns the robot smoothly to the nominal route \cite{losey2017trajectory}. The physical displacement produced while yielding is excluded from the commanded trajectory deformation.

\p{Dependent Variables}
Clearance is the fraction of trials that avoid the obstacle. Joint success additionally requires goal completion, payload retention, and no termination. Goal error measures the final absolute lateral distance from the goal.

\p{Results}
\textbf{Oracle} establishes a $92.4\%$ true-force performance ceiling, while \textbf{Ours} achieves $77.1\%$ clearance and $76.9\%$ joint success using only onboard measurements. \textbf{Ours-Analytic} clears $34.8\%$ of trials, and \textsc{Facet-L} cannot deliberately alter its route. The small gap between clearance and joint success shows that successful avoidance generally also preserves the payload and reaches the goal, demonstrating that the learned force estimate is useful beyond the immediate compliant response.

\begin{table}[t]
    \centering
    \caption{Force-guided obstacle-avoidance performance. Values in the columns show the mean and std. dev. across five seeds.}
    \vspace{-0.5em}
    \label{tab:deflection_clearance}
    \scriptsize
    \setlength{\tabcolsep}{5pt}
    \begin{tabular}{@{}l c c c@{}}
        \toprule
        Method
        & \shortstack{Clearance\\(\%) $\uparrow$}
        & \shortstack{Joint success\\(\%) $\uparrow$}
        & \shortstack{Goal error\\(m) $\downarrow$} \\
        \midrule
        Oracle
            & $92.4\pm0.8$ & $92.4\pm0.7$ & $0.038\pm0.002$ \\
        \textsc{Facet-L}
            & $0.0\pm0.0$ & $0.0\pm0.0$ & $0.093\pm0.002$ \\
        Ours-Analytic
            & $34.8\pm1.8$ & $34.3\pm1.7$ & $0.055\pm0.003$ \\
        Ours
            & $77.1\pm1.6$ & $76.9\pm1.5$ & $0.068\pm0.002$ \\
        \bottomrule
    \end{tabular}
    \vspace{-2.0em}
\end{table}

\section{Real-World Experiments} \label{experiments}

Now that we have validated our proposed approach for external force-aware compliant locomotion in simulations, we move to real world experiments to evaluate the performance of our approach. We evaluate the approach in nominal path following and human-guided route correction, allowing us to examine whether the force-aware response preserves ordinary locomotion while providing a useful interface when an interaction is applied to the payload.

\p{Experimental Setup}
We deploy the policies on a Unitree Go2 carrying an unrestrained $4$~kg payload. The impedance parameters for the controller are fixed at $K_p=8$~N/m, $K_d=1.8\sqrt{K_p}$~N\,s/m, and $M_v=4$~kg. An external motion capture system provides the route-relative position and heading used by the path follower and records the executed trajectory. Motion capture data is not provided as an input to the policy or force estimator, but only used as an external measurement system to track the robot.

\p{Independent Variables}
We compare \textbf{Ours} with the task-matched {\textsc{Facet}-L baseline introduced in Section~\ref{sec:c1c2}. Both methods track a constant nominal speed of $0.25$m/s and cover a total forward distance of $6$m from the initial position.

\begin{table*}[t]
    \centering
    \caption{Quantitative results for real-world straight-line walking and human-guided trajectory correction experiments. Values in the columns represent the mean and std. dev. across ten trials.}
    \vspace{-0.5em}
    \label{tab:hardware_results}
    \scriptsize
    \setlength{\tabcolsep}{6.5pt}
    \begin{tabular}{@{}l l c c c c c c c c@{}}
        \toprule
        & Method
        & \shortstack{$v_x$\\(m/s)}
        & \shortstack{$|v_y|$\\(m/s)}
        & \shortstack{$v_{xy}$\\(m/s)}
        & \shortstack{Mean path\\deviation (m) $\downarrow$}
        & \shortstack{Goal \\error (m) $\downarrow$}
        & \shortstack{Mean tilt\\(deg) $\downarrow$}
        & \shortstack{Peak tilt\\(deg) $\downarrow$} \\
        \midrule
        \multirow{2}{3.0em}{Straight Line}
        & \textsc{Facet}-L
            & $0.259\pm0.001$ & $0.001\pm0.001$ & $0.259\pm0.001$
            & $0.025\pm0.017$ & $0.026\pm0.025$
            & $2.05\pm0.43$ & $6.14\pm1.91$ \\
        & Ours
            & $0.252\pm0.002$ & $0.003\pm0.003$ & $0.252\pm0.002$
            & $0.045\pm0.020$ & $0.042\pm0.015$
            & $1.70\pm0.82$ & $7.40\pm5.01$ \\
        \midrule
        \multirow{2}{3.0em}{Human Guided}
        & \textsc{Facet}-L
            & $0.232\pm0.006$ & $0.109\pm0.023$ & $0.258\pm0.009$
            & $0.058\pm0.017$ & $0.089\pm0.025$
            & $2.07\pm0.43$ & $8.38\pm2.13$ \\
        & Ours
            & $0.195\pm0.012$ & $0.160\pm0.032$ & $0.254\pm0.018$
            & $0.096\pm0.015$ & $0.045\pm0.045$
            & $3.45\pm0.53$ & $13.37\pm4.62$ \\
        \bottomrule
    \end{tabular}
    \vspace{-1.5em}
\end{table*}

\p{Procedure}
During straight-line walking, the robot follows the same fixed route throughout the trial. The estimated force remains available to our low-level policy and compliant reference, allowing the controller to respond to motion of the unrestrained payload.

During the human-guided trajectory correction, we use a minimum interaction duration of 1 second to begin replanning.
When an external interaction is detected and persists beyond this threshold, \textbf{Ours} pauses forward progression, yields to the push, and computes the new path from the displaced position. The accumulated impulse determines the deformation in the trajectory, similar to Section \ref{sec:simulation_deflection}. In contrast, \textsc{Facet-L}, which does not have an explicit force estimate, recomputes a straight line path from the displaced position to the goal. Because the methods generate different post-interaction trajectories, this experiment evaluates their end-to-end task behavior rather than directly comparing trajectory-generation or path-tracking accuracy. \textsc{Facet-L} serves as a force-unaware baseline that resumes toward the goal from its displaced pose, whereas \textbf{Ours} uses the estimated interaction to produce a deliberate route correction.

\p{Dependent Variables}
In real-world, we do not have access to ground-truth external forces, we only evaluate the resulting motion of the robot. Additionally, since \textsc{Facet-L} cannot explicitly estimate external forces, but can only use implicit measurements to react to them, we evaluate the resulting robot motion based on the respective nominal and deformed trajectories of each method.
To evaluate the velocity tracking behavior, we measure the average forward and lateral velocities, and the average speed across each trajectory. Across both experiments, an average speed closer to the commanded velocity indicates a better performance. 
In addition to velocity, we also measure mean deviation from the commanded path, final error from the goal location, and mean and peak tilt of the robot when carrying the payload.

\begin{figure}[t]
    \centering
    \includegraphics[width=1.0\columnwidth]{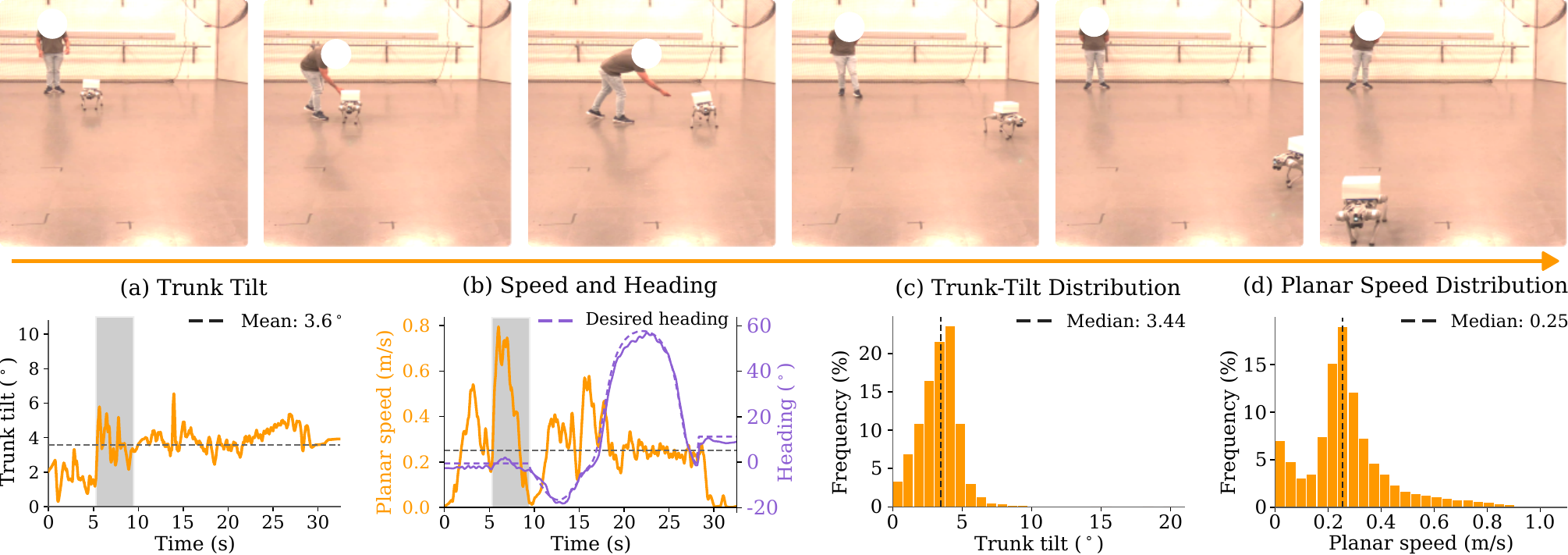}
		\vspace{-1.5em}
    \caption{8, Real world human-guided trajectory correction using our proposed approach. The top row shows a representative interaction sequence. The bottom row reports (a) trunk tilt and (b) planar speed together with the desired and measured headings for the same run. Gray shading marks the interaction, yielding, and replanning interval. Panels (c) and (d) show trunk-tilt and planar-speed histograms across ten trials. 
    }
    \label{fig:hardware_response}
		\vspace{-1.5em}
\end{figure}

\p{Results}
The results for this experiment are summarized in Table \ref{tab:hardware_results} and Figure \ref{fig:hardware_response}. We first start by analyzing the straight-line walking behavior of the robot. We observe that \textbf{Ours} closely tracks the commanded walking speed, while maintaining a near-zero lateral speed. We also observe that \textsc{Facet-L} follows the straight route more tightly, with a mean lateral deviation of $0.025\pm0.017$~m compared with $0.045\pm0.020$~m for \textbf{Ours}. Although \textbf{Ours} has the lower mean tilt, we hypothesize that its larger mean path deviation and higher peak tilt are a result of corrective motions of the robot caused by the force estimator detecting payload movement when transporting an unrestrained payload.

Next, we analyze the behavior of the robot under an external interaction. We observe that both approaches have a higher lateral velocity and lower forward velocity as compared to the straight-line walking, reflecting the intended yielding and route correction behaviors. Here, \textbf{Ours} has a higher lateral velocity as compared to \textsc{Facet-L}, showing that \textbf{Ours} yields more compliantly to external interactions, consistent with our simulation results in Table \ref{tab:simulation_compliance_aggregate}. 
In the representative trial in Figure \ref{fig:hardware_response}, the robot speed reduces during the interaction interval and the robot resumes its motion after a new path is available. Across all ten trials, the median trunk tilt is $3.44^\circ$ and the median speed is $0.25$~m/s.
We observe that the route correction leads to an average lateral displacement of $0.345\pm0.081$m beyond the displaced pose, and \textbf{Ours} reaches the goal more accurately as compared to \textsc{Facet-L}. 
Together, these results show that our approach preserves stable nominal locomotion while enabling payload interactions to produce purposeful, force-guided changes in the robot’s trajectory.

\section{Conclusion}
In this paper, we presented a force-aware locomotion framework for quadruped robots carrying unrestrained payloads. Instead of assuming that external interactions act directly on the robot or through a rigidly connected load, our method considers forces applied directly to a payload coupled to the robot only through frictional contact. We separate the learning of force-aware locomotion and force estimation, first learning the desired compliant load-carrying behavior and then training a causal estimator to recover payload interactions from onboard robot measurements. Finally, we finetune the locomotion policy with the learned estimator in the loop to account for errors introduced during deployment. Our simulation and real-world experiments suggest that the resulting controller can maintain stable load-carrying locomotion while yielding compliantly to payload interactions. We further show that the estimated force provides a useful interface for higher-level behavior, enabling a human to guide the robot’s trajectory by interacting directly with the payload.

\p{Limitations}
Our current formulation considers planar payload interactions and assumes that the payload remains in contact with its support. However, sufficiently large forces may still cause payload loss. Future work will extend the approach to three-dimensional interactions and a broader range of payloads and terrains.


\balance
\bibliographystyle{IEEEtran}
\bibliography{IEEEabrv,bibtex}

\end{document}